\documentclass[aps,pre,groupedaddress,twocolumn]{revtex4}

\usepackage{amsmath}
\usepackage{amssymb}
\usepackage{graphicx}
\usepackage{color}
\usepackage{bm}
\usepackage{subfigure}
\usepackage{multirow}
\usepackage{latexsym}

\begin{document}
\title{Variational Neural Networks: Every Layer and Neuron Can Be Unique}
\author{Yiwei Li \\ \href{mailto: mrhutumeng@gmail.com}{mrhutumeng@gmail.com} \\ Enzhi Li \\ \href{mailto: enzhililsu@gmail.com}{enzhililsu@gmail.com}}
\begin{abstract}
The choice of activation function can significantly influence the performance of neural networks. The lack of guiding principles for the selection of activation function is lamentable. We try to address this issue by introducing our variational neural networks, where the activation function is represented as a linear combination of possible candidate functions, and an optimal activation is obtained via minimization of a loss function using gradient descent method. The gradient formulae for the loss function with respect to these expansion coefficients are central for the implementation of gradient descent algorithm, and here we derive these gradient formulae. 
\end{abstract}

\maketitle

\section{Introduction}
\label{secIntro}

In conventational artificial neural networks (ANNs), the backward-propagation updates weights for the entire network to minimize the loss function\cite{lecun1998gradient}. 
The activation function in each hidden layer is determined before training the network and fixed during the training process. Various activation functions have been proposed, such as sigmoid, $\tanh$, ReLu, etc \cite{goodfellow2016deep}. There may be other customized activation functions for specific use cases. Empirically, ReLu is the default choice when building deep neural networks for computer vision and speech recognition\cite{lecun2015deep}. 

The choice of activation function is pretty arbitrary. During the early development stage, the practitioners of neural networks tried to simulate genuine neurons in human, and preferred to use activation functions that saturate when its input value is large. Moreover, previously, people have held it to be self evident that the activation functions should be differentiable everywhere, and thus sigmoid and hyperbolic tangent functions are widely used. The realization that ReLu, which is neither saturating nor everywhere differentiable, could also be an activation function with even better performance than sigmoid or hyperbolic tangent removes the shackle in people's imagination, and significantly promotes the development of neural networks\cite{glorot2011deep}.  However, up to now, the choice of activation functions is still ad-hoc and no rigorous proof exists that can demonstrate that ReLU or any other novel activation function is superior to the conventional sigmoid and hyperbolic tangent functions. The advantage of one activation function over another is established from mere experience, and the lack of theoretical or algorithmic justification thereof is one of our concerns. 

Here, in this article, we propose a more systematic method to enable our neural network to find the optimal activation function automatically. In order to do this, we first select a set of candidate eigen functions, and then make the assumption that the optimal activation function can be represented as as linear combination of these candidate eigen functions, the combination coefficients of which yet to be determined. We next define a loss function, and try to minimize this loss function with respect to the combination coefficients, together with the weight matrices and biases that are used in conventional artificial neural networks. The activation function, which is now a linear combination of basis eigen functions, can be unique for each hidden layer (even the output layer) or each neuron in each hidden layer (even the output layer). Since in our system, the activation functions are determined from variational principle, we name our neural network as variational neural network (VNN). In this article, we will derive the formulae for the gradients of loss function with respect to the expansion coefficients. Programmatic implementation of this algorithm is still in its way. We find that in our variational neural network, back propagation method still applies, and thus no major modification of conventional neural network programs is needed. 

The organization of this paper is as follows. In section \ref{layer_level_unique}, we derive the gradient formulae for loss function with respect to the expansion coefficients. In this section, the activation functions are represented as a linear combination of candidate eigen functions, and we impose a restraint that neurons residing in the same layer should share the same activation function. In section \ref{neuron_level_unique}, we relax the restraint in section \ref{layer_level_unique}, and now each neuron has its own unique activation function. The gradient formulae for the loss function with respect to the expansion coefficients are derived in a similar manner as that in section \ref{layer_level_unique}. We make a conclusion in section \ref{conclusion}. 

\section{Variation of Hidden Layer Activation Functions}
\label{layer_level_unique}
Suppose $E$ is the loss function, $\sigma$ is the activation function in output layer, $\omega_{ij}^{(N)}$ is the weight on the connection between the 
$i^{\textrm{th}}$ nueron in $(N-1)^{\textrm{th}}$ hidden layer and the $j^{\textrm{th}}$ neuron in $N^{\textrm{th}}$ hidden layer, $\omega_{ij}^{(O)}$ is 
the weight on the connection between the $i^{\textrm{th}}$ neuron in the last hidden layer and the $j^{\textrm{th}}$ neuron in the output layer, and $\mathcal{F}^{(N)}$ 
is the activation function in $N^{\textrm{th}}$ hidden layer. $\mathcal{F}^{(N)}(x) = \sum\limits_{i=1}^{\infty}\alpha_i^{(N)} f_i (x)$, where $f_i(x)$ is 
a set of eigen-functions (for example, it can be $\sin(i\omega x)$ and $\cos(i\omega x)$). In practice, we can cut-off the summation to a large number $M$: 
$\mathcal{F}^{(N)}(x) \approx \sum\limits_{i=1}^{M}\alpha_i^{(N)} f_i (x)$.

Now during the backward-propagation, the network will not only update the weights in each connection between neurons, but also update the weights of the
eigen-functions in activation function. The neuron connection weight update is the same as before. The eigen function weight update for the last hidden layer 
is as follows (suppose we have $N$ hidden layers in total),
\begin{align} \label{eq:weight-update}
 \alpha_i^{(N)} = \alpha_i^{(N)} - \eta\frac{\partial E}{\partial \alpha_i^{(N)}} \,,
\end{align}
where $\eta$ is the learning rate. Using the chain rule, we get (suppose the loss function has such format $E = \sum\limits_{l=1}^{n}\mathcal{E}(t_l, \sigma(\textrm{net}_l^{(O)}))$, 
where $t_l$ is the $l^{\textrm{th}}$ element of the label)
\begin{align} \label{eq:alpha}
 \frac{\partial E}{\partial \alpha_i^{(N)}} = \sum\limits_{l=1}^{n} \frac{\partial \mathcal{E}}{\partial \sigma} \cdot \frac{\partial \sigma}{\partial \textrm{net}_l^{(O)}} \cdot \frac{\partial \textrm{net}_l^{(O)}}{\partial \alpha_i^{(N)}} \,,
\end{align}
where $\frac{\partial E}{\partial \sigma}$ is based on the format of loss function and $\frac{\partial \sigma}{\partial \textrm{net}_l^{(O)}}$ is based on the format of 
activation function in the output layer, which are straight-forward to compute, $\textrm{net}_l^{(O)}$ is the input from the $l^{\textrm{th}}$ neuron in the 
output layer. Suppose there are $m$ neurons in the last hidden layer
\begin{align} \label{eq:net}
 \textrm{net}_l^{(O)} &= \sum\limits_{j=1}^{m} \omega_{jl}^{(O)} \mathcal{F}^{(N)}(\textrm{net}_j^{(N)}) \nonumber \\
		&= \sum\limits_{j=1}^{m} \omega_{jl}^{(O)} \sum\limits_{k=1}^{M} \alpha_k^{(N)} f_k (\textrm{net}_j^{(N)}) \,.
\end{align}
The last part of Eq.~(\ref{eq:alpha}) is
\begin{align} \label{eq:alpha2}
 \frac{\partial \textrm{net}_l^{(O)}}{\partial \alpha_i^{(N)}} = \sum\limits_{j=1}^{m} \omega_{jl}^{(O)} f_i(\text{net}_{j}^{(N)}) \,.
\end{align}
Therefore, the eigen function weight update for the last hidden layer is
\begin{align} \label{eq:last_hidden_layer}
 \frac{\partial E}{\partial \alpha_i^{(N)}} = \sum\limits_{l=1}^{n} \frac{\partial \mathcal{E}}{\partial \sigma} \cdot \frac{\partial \sigma}{\partial \textrm{net}_l^{(O)}} \cdot \sum\limits_{j=1}^{m} \omega_{jl}^{(O)} f_i(\text{net}_{j}^{(N)}) \,.
\end{align}
If we transform Eq.~(\ref{eq:last_hidden_layer}) into matrix format, then
\begin{align} \label{eq:matrix1}
 \frac{\partial E}{\partial \alpha_i^{(N)}} = \left( \overrightarrow{f}_i^{(N)} \right)^{\textrm{T}} W^{(O)} \nabla_{\overrightarrow{\textrm{net}}^{(O)}} \mathcal{E} \,,
\end{align}
where $(\nabla_{\overrightarrow{\textrm{net}}^{(O)}} \mathcal{E})_l = \frac{\partial \mathcal{E}}{\partial \sigma} \cdot \frac{\partial \sigma}{\partial \textrm{net}_l^{(O)}}$, 
$W^{(O)}$ is a $m\times n$ matrix [$(W^{(O)})_{jl} = \omega_{jl}^{(O)}$], $\textrm{T}$ is transpose operation, and 
$\left( \overrightarrow{f}_i^{(N)} \right)^{\textrm{T}} = \left[ f_i(\textrm{net}_1^{(N)}),\dots,f_i(\textrm{net}_n^{(N)}) \right]$ .

Similarly, the eigen function weight update for the second last hidden layer is
\begin{align} \label{eq:second_alpha}
 \frac{\partial E}{\partial \alpha_i^{(N-1)}} = \sum\limits_{l=1}^{n} \frac{\partial \mathcal{E}}{\partial \sigma} \cdot \frac{\partial \sigma}{\partial \textrm{net}_l^{(O)}} \cdot \frac{\partial \textrm{net}_l^{(O)}}{\partial \alpha_i^{(N-1)}} \,,
\end{align}
where
\begin{eqnarray} \label{eq:second_net}
 && \textrm{net}_l^{(O)} = \sum\limits_{j=1}^{m} \omega_{jl}^{(O)} \sum\limits_{k=1}^{M} \alpha_k^{(N)} f_k(\textrm{net}_j^{(N)}) \nonumber \\
 && \textrm{net}_j^{(N)} = \sum\limits_{p=1}^{s} \omega_{pj}^{(N)} \sum\limits_{q=1}^{M} \alpha_q^{(N-1)} f_q(\textrm{net}_p^{(N-1)}) \,.
\end{eqnarray}
Therefore, the eigen function weight update is
\begin{align} \label{eq:second_last_hidden_layer}
 \frac{\partial E}{\partial \alpha_i^{(N-1)}} =& \sum\limits_{l=1}^{n} \frac{\partial \mathcal{E}}{\partial \sigma} \cdot \frac{\partial \sigma}{\partial \textrm{net}_l^{(O)}} \cdot \sum\limits_{j=1}^{m} \omega_{jl}^{(O)} \nonumber \\ 
					       & \cdot \sum\limits_{k=1}^{M} \alpha_k^{(N)} \frac{\partial f_k}{\partial \textrm{net}_j^{(N)}} \sum\limits_{p=1}^{s} \omega_{pj}^{(N)} f_i(\text{net}_{p}^{(N-1)}) \,.
\end{align}
If we transform Eq.~(\ref{eq:second_last_hidden_layer}) into matrix format, then
\begin{align} \label{eq:matrix2}
 \frac{\partial E}{\partial \alpha_i^{(N-1)}} = \left( \overrightarrow{f}_i^{(N-1)} \right)^{\textrm{T}} W^{(N)} \widetilde{W}^{(O)} \nabla_{\overrightarrow{\textrm{net}}^{(O)}} \mathcal{E} \,,
\end{align}
where $\left[ \widetilde{W}^{(O)} \right]_{ij} = \omega_{ij}^{(O)} \cdot \frac{\partial \mathcal{F}^{(N)}}{\partial \textrm{net}_i^{(N)}}$.

It is easy to demonstrate that the general formula of updating the $i^{\textrm{th}}$ eigen function weight for the $\gamma^{\textrm{th}}$ last hidden layer
\begin{align} \label{eq:matrix_general}
 \frac{\partial E}{\partial \alpha_i^{(N-\gamma+1)}} =& \left( \overrightarrow{f}_i^{(N-\gamma+1)} \right)^{\textrm{T}} W^{(N-\gamma+2)} \nonumber \\ 
						    & \cdot \widetilde{W}^{(N-\gamma+3)} \cdots \widetilde{W}^{(N)} \widetilde{W}^{(O)} \nabla_{\overrightarrow{\textrm{net}}^{(O)}} \mathcal{E} \,,
\end{align}
where $\left[ \widetilde{W}^{(\beta)} \right]_{ij} = \omega_{ij}^{(\beta)} \cdot \frac{\partial \mathcal{F}^{(\beta-1)}}{\partial \textrm{net}_i^{(\beta-1)}}$.

Therefore, the entire update of the eigen function weights for the $\gamma^{\textrm{th}}$ last hidden layer is
\begin{align} \label{eq:matrix_general2}
 \nabla_{\overrightarrow{\alpha}^{(N-\gamma+1)}} E =& \left( \frac{\partial E}{\partial \alpha_1^{(N-\gamma+1)}}, \dots, \frac{\partial E}{\partial \alpha_M^{(N-\gamma+1)}} \right)^{\textrm{T}} \nonumber \\
						   =& \overrightarrow{F}^{(N-\gamma+1)} W^{(N-\gamma+2)} \nonumber \\ 
						    & \cdot \widetilde{W}^{(N-\gamma+3)} \cdots \widetilde{W}^{(N)} \widetilde{W}^{(O)} \nabla_{\overrightarrow{\textrm{net}}^{(O)}} \mathcal{E} \,,
\end{align}
where $\left[ \overrightarrow{F}^{(N-\gamma+1)} \right]_{ij} = f_i(\textrm{net}_j^{(N-\gamma+1)})$.

In general, we can also treat the activation function of the output layer as a summation of $M$ eigen functions with distinct weights. In this case, we can regard $\sigma$ as a scaling function (e.g. softmax), then 
the format of the formula above would remain the same.

\section{Variation of Hidden Node Activation Functions}
\label{neuron_level_unique}
Theoretically, there is no constraint that all the nodes in the same hidden layer must have exactly same activation function. So we can generalize our method such that 
each node in each hidden layer can have its unique activation function $\mathcal{F}_j^{(N)}$ (the activation function for the $j^{\textrm{th}}$ node in the $N^{\textrm{th}}$ 
hidden layer), where $\mathcal{F}_j^{(N)} \approx \sum\limits_{i=1}^{M}\alpha_{j,i}^{(N)} f_i (x) $. Then, the update of the $i^{\textrm{th}}$ eigen function weight in 
the $j^{\textrm{th}}$ neuron of the last hidden layer is
\begin{align} \label{eq:neuron_update_last_layer}
 \frac{\partial E}{\partial \alpha_{j,i}^{(N)}} = \sum\limits_{l=1}^n \frac{\partial \mathcal{E}}{\partial \textrm{net}_l^{(O)}} \cdot \frac{\partial \textrm{net}_l^{(O)}}{\partial \alpha_{j,i}^{(N)}} \,,
\end{align}
where 
\begin{align}
\textrm{net}_l^{(O)} = \sum\limits_{k=1}^m \omega_{kl}^{(O)} \mathcal{F}_k^{(N)}(\textrm{net}_k^{(N)}) = \sum\limits_{k=1}^{m} \omega_{kl}^{(O)} \sum\limits_{a=1}^M \alpha_{k,a}^{(N)} f_a(\textrm{net}_k^{(N)}) \nonumber \,.
\end{align}
Then, we get
\begin{align} \label{eq:neuron_update_last_layer2}
 \frac{\partial E}{\partial \alpha_{j,i}^{(N)}} = \sum\limits_{l=1}^n \frac{\partial \mathcal{E}}{\partial \textrm{net}_l^{(O)}} \cdot \omega_{jl}^{(O)} f_i(\textrm{net}_j^{(N)}) \,.
\end{align}
So the update of all the eigen function weights in the last hidden layer is
\begin{align} \label{eq:matrix_last_layer}
 \Delta_{\alpha^{(N)}} E = F^{(N)} \odot \left( W^{(O)} \nabla_{\overrightarrow{\textrm{net}}^{(O)}} \mathcal{E} \right) \,,
\end{align}
where $\left[ \Delta_{\alpha^{(N)}} E \right]_{ij} = \frac{\partial E}{\partial \alpha_{j,i}^{(N)}}$, $\left[ F^{(N)} \right]_{ij} = f_i(\textrm{net}_j^{(N)})$, and 
the operation $\odot$ is the element-wise multiplication between each column of $F^{(N)}$ and the vector $\left( W^{(O)} \nabla_{\overrightarrow{\textrm{net}}^{(O)}} \mathcal{E} \right)$.

Similarly, the update of the $i^{\textrm{th}}$ eigen function weight in the $j^{\textrm{th}}$ neuron of the second last hidden layer is
\begin{align} \label{eq:neuron_update_second_last_layer}
 \frac{\partial E}{\partial \alpha_{j,i}^{(N-1)}} = \sum\limits_{l=1}^n \frac{\partial \mathcal{E}}{\partial \textrm{net}_l^{(O)}} \cdot \frac{\partial \textrm{net}_l^{(O)}}{\partial \alpha_{j,i}^{(N-1)}} \,,
\end{align}
where
\begin{eqnarray} 
 && \textrm{net}_l^{(O)} = \sum\limits_{k=1}^m \omega_{kl}^{(O)} \mathcal{F}_k^{(N)}(\textrm{net}_k^{(N)}) \nonumber \\
 && \textrm{net}_v^{(N)} = \sum\limits_{a=1}^s \omega_{av}^{(N)} \sum\limits_{b=1} \alpha_{a,b}^{(N-1)} f_b(\textrm{net}_a^{(N-1)}) \,.
\end{eqnarray}
Then, we get
\begin{align} \label{eq:neuron_update_last_layer2}
 \frac{\partial E}{\partial \alpha_{j,i}^{(N-1)}} = \sum\limits_{l=1}^n \frac{\partial \mathcal{E}}{\partial \textrm{net}_l^{(O)}} \sum\limits_{k=1}^m \omega_{kl}^{(O)} \frac{\partial \mathcal{F}_{k}^{(N)}}{\partial \textrm{net}_k^{(N)}} \omega_{jk}^{(N)} f_i(\textrm{net}_j^{(N-1)}) \,.
\end{align}
Then, the update of all the eigen function weights in the second last hidden layer is
\begin{align} \label{eq:matrix_second_last_layer}
 \Delta_{\alpha^{(N-1)}} E = F^{(N-1)} \odot \left( W^{(N)} \nabla_{\overrightarrow{\textrm{net}}^{(N)}} \mathcal{F}^{(N)} \odot \left( W^{(O)} \nabla_{\overrightarrow{\textrm{net}}^{(O)}} \mathcal{E} \right) \right) \,,
\end{align}
where $\left[ \nabla_{\overrightarrow{\textrm{net}}^{(N)}} \mathcal{F}^{(N)} \right]_k = \frac{\partial \mathcal{F}_{k}^{(N)}}{\partial \textrm{net}_k^{(N)}}$ and 
$\left[ F^{(N-1)} \right]_{ij} = f_i(\textrm{net}_j^{(N-1)})$.

Therefore, the update of all the eigen function weights in the $(\gamma+1)^{\textrm{th}}$ last hidden layer is
\begin{align} \label{eq:matrix_final}
 \Delta_{\overrightarrow{\alpha}^{(N-\gamma)}} E =& F^{(N-\gamma)} \odot ( W^{(N-\gamma+1)} \nabla_{\overrightarrow{\textrm{net}}^{(N-\gamma+1)}} \mathcal{F}^{(N-\gamma+1)} \nonumber \\
						  & \odot ( \cdots \odot ( W^{(N)} \nabla_{\overrightarrow{\textrm{net}}^{(N)}} \mathcal{F}^{(N)} \nonumber \\ 
						  & \odot (W^{(O)} \nabla_{\overrightarrow{\textrm{net}}^{(O)}} \mathcal{E})) \cdots )) \,.
\end{align}

In general, we can also treat the activation function of the neurons in the output layer as a summation of $M$ eigen functions with distinct weights. The formula format will remain the same. 
However, the problem is that, take classification as an instance, the outputs of neurons in the output layer should be probability or probability-like values. If we use different activation 
function for different output neurons, it is very difficult to tell what is the meaning of the outcomes from the output layer. Thus, here, we choose not to vary the activation functions for the output layer. 

\section{Conclusion and outlook}
\label{conclusion}
In this article, we have proposed a method to allow each layer and even each neuron in the neural networks to have its own activation function. The activation functions are represented as a linear combination of basis eigen functions. We train the neural network by minimizing a loss function with respect to these expansion coefficients together with conventional weight matrices and biases. After training the networks, we will not only be able to get optimal weights and biases between neurons in nearest layers, but also the optimal activation functions. Our ongoing work will focus on building a real model and test the performance of this variational neural networks against conventional neural networks.

\bibliography{ref}

\begin{thebibliography}{4}
\expandafter\ifx\csname natexlab\endcsname\relax\def\natexlab#1{#1}\fi
\expandafter\ifx\csname bibnamefont\endcsname\relax
  \def\bibnamefont#1{#1}\fi
\expandafter\ifx\csname bibfnamefont\endcsname\relax
  \def\bibfnamefont#1{#1}\fi
\expandafter\ifx\csname citenamefont\endcsname\relax
  \def\citenamefont#1{#1}\fi
\expandafter\ifx\csname url\endcsname\relax
  \def\url#1{\texttt{#1}}\fi
\expandafter\ifx\csname urlprefix\endcsname\relax\def\urlprefix{URL }\fi
\providecommand{\bibinfo}[2]{#2}
\providecommand{\eprint}[2][]{\url{#2}}

\bibitem[{\citenamefont{LeCun et~al.}(1998)\citenamefont{LeCun, Bottou, Bengio,
  and Haffner}}]{lecun1998gradient}
\bibinfo{author}{\bibfnamefont{Y.}~\bibnamefont{LeCun}},
  \bibinfo{author}{\bibfnamefont{L.}~\bibnamefont{Bottou}},
  \bibinfo{author}{\bibfnamefont{Y.}~\bibnamefont{Bengio}}, \bibnamefont{and}
  \bibinfo{author}{\bibfnamefont{P.}~\bibnamefont{Haffner}},
  \bibinfo{journal}{Proceedings of the IEEE} \textbf{\bibinfo{volume}{86}},
  \bibinfo{pages}{2278} (\bibinfo{year}{1998}).

\bibitem[{\citenamefont{Goodfellow et~al.}(2016)\citenamefont{Goodfellow,
  Bengio, Courville, and Bengio}}]{goodfellow2016deep}
\bibinfo{author}{\bibfnamefont{I.}~\bibnamefont{Goodfellow}},
  \bibinfo{author}{\bibfnamefont{Y.}~\bibnamefont{Bengio}},
  \bibinfo{author}{\bibfnamefont{A.}~\bibnamefont{Courville}},
  \bibnamefont{and} \bibinfo{author}{\bibfnamefont{Y.}~\bibnamefont{Bengio}},
  \emph{\bibinfo{title}{Deep learning}}, vol.~\bibinfo{volume}{1}
  (\bibinfo{publisher}{MIT press Cambridge}, \bibinfo{year}{2016}).

\bibitem[{\citenamefont{LeCun et~al.}(2015)\citenamefont{LeCun, Bengio, and
  Hinton}}]{lecun2015deep}
\bibinfo{author}{\bibfnamefont{Y.}~\bibnamefont{LeCun}},
  \bibinfo{author}{\bibfnamefont{Y.}~\bibnamefont{Bengio}}, \bibnamefont{and}
  \bibinfo{author}{\bibfnamefont{G.}~\bibnamefont{Hinton}},
  \bibinfo{journal}{nature} \textbf{\bibinfo{volume}{521}},
  \bibinfo{pages}{436} (\bibinfo{year}{2015}).

\bibitem[{\citenamefont{Glorot et~al.}(2011)\citenamefont{Glorot, Bordes, and
  Bengio}}]{glorot2011deep}
\bibinfo{author}{\bibfnamefont{X.}~\bibnamefont{Glorot}},
  \bibinfo{author}{\bibfnamefont{A.}~\bibnamefont{Bordes}}, \bibnamefont{and}
  \bibinfo{author}{\bibfnamefont{Y.}~\bibnamefont{Bengio}}, in
  \emph{\bibinfo{booktitle}{Proceedings of the fourteenth international
  conference on artificial intelligence and statistics}}
  (\bibinfo{year}{2011}), pp. \bibinfo{pages}{315--323}.

\end{thebibliography}

\end{document}